\documentclass{article}
\usepackage{graphicx}
\usepackage{hyperref}
\usepackage{booktabs}
\usepackage{float}
\usepackage{multirow}
\usepackage{caption}
\usepackage{adjustbox}
\usepackage[preprint]{neurips_2025}
\usepackage{amsmath}

\title{Self-Supervised YOLO: Leveraging Contrastive Learning for Label-Efficient Object Detection}

\author{%
  Manikanta Kotthapalli
  \And
  Reshma Bhatia
  \And
  Nainsi Jain
}

\begin{document}
\maketitle

\begin{abstract}\label{sec:abstr}
One-stage object detectors such as the YOLO family achieve state-of-the-art performance in real-time vision applications but remain heavily reliant on large-scale labeled datasets for training. In this work, we present a systematic study of contrastive self-supervised learning (SSL) as a means to reduce this dependency by pretraining YOLOv5 and YOLOv8 backbones on unlabeled images using the SimCLR framework. Our approach introduces a simple yet effective pipeline that adapts YOLO’s convolutional backbones as encoders, employs global pooling and projection heads, and optimizes a contrastive loss using augmentations of the COCO unlabeled dataset (120k images). The pretrained backbones are then fine-tuned on a cyclist detection task with limited labeled data \cite{li2016cyclist}. Experimental results show that SSL pretraining leads to consistently higher mAP, faster convergence, and improved precision-recall performance, especially in low-label regimes. For example, our SimCLR-pretrained YOLOv8 achieves a mAP@50:95 of 0.7663, outperforming its supervised counterpart despite using no annotations during pretraining. These findings establish a strong baseline for applying contrastive SSL to one-stage detectors and highlight the potential of unlabeled data as a scalable resource for label-efficient object detection.
\end{abstract}

\section{Introduction}\label{sec:intro}
Object detection models have achieved remarkable success by leveraging large annotated datasets. In particular, one-stage detectors like the \textbf{YOLO} family \cite{redmon2016yolo, bochkovskiy2020yolov4} are popular for real-time applications due to their speed and accuracy. However, training these models still heavily relies on supervised pretraining on massive labeled datasets (e.g. ImageNet or COCO), which incurs high human labeling cost. Reducing the dependence on labeled data is crucial for scaling object detection to new domains and tasks where annotations are scarce or expensive to obtain. \textbf{Self-supervised learning (SSL)} has emerged as a promising paradigm to utilize vast amounts of \emph{unlabeled} data to pretrain deep neural networks \cite{chen2020simple,he2020moco,grill2020byol}. By learning general feature representations without human annotations, SSL can potentially provide a strong initialization for detection models, thereby minimizing the need for manually labeled images.

Recent advances in SSL for computer vision have significantly closed the gap with supervised pretraining on downstream tasks. Methods based on \textit{contrastive learning} \cite{chen2020simple,he2020moco,grill2020byol} train an encoder to discriminate between different images or views, yielding transferable representations that improve object detection and segmentation after fine-tuning \cite{he2020moco,wang2021densecl}. More recently, approaches like \textit{masked image modeling} \cite{he2022mae,Xie_2022_CVPR} and \textit{self-distillation with no labels} \cite{caron2021dino,oquab2023dinov2} have produced even stronger visual features. For example, Masked Autoencoders (MAE) \cite{he2022mae} pretrain vision transformers by reconstructing masked patches and have been shown to outperform supervised ImageNet pretraining on COCO object detection \cite{he2022mae}. Similarly, the DINO framework \cite{caron2021dino} and its recent extension DINOv2 \cite{oquab2023dinov2} use teacher-student networks to learn high-level semantic features that transfer robustly to detection and segmentation tasks. These developments suggest that modern SSL methods can provide powerful initialization for detection models, potentially reducing the required amount of labeled data.

Despite this progress, most prior work on SSL for vision has focused on classification backbones (e.g. ResNet or ViT) and two-stage detectors. \textbf{One-stage detectors like YOLO remain under-explored in the SSL context}. YOLO architectures differ from standard classification networks due to their multi-scale feature maps and detection-specific heads, which raises the question of how to effectively pretrain them with SSL. To our knowledge, there have been only limited attempts to apply self-supervised pretraining to YOLO-style models. In this paper, we address this gap by systematically studying SSL pretraining for YOLO detectors. We specifically target the popular YOLOv5 \cite{jocher2020yolov5} and the newer YOLOv8 \cite{yolov8} architectures, examining how SSL can benefit each.

Our core idea is to pretrain the \emph{backbone} of YOLO on unlabeled data using a contrastive SSL objective, and then fine-tune the full detector on a target task with limited labeled examples. We use SimCLR \cite{chen2020simple} as a representative contrastive method for pretraining. We leverage the large unlabeled portion of the MS~COCO dataset to pretrain YOLO backbones without any human labels. We then fine-tune the pretrained YOLOv5 \cite{jocher2020yolov5} and YOLOv8 \cite{yolov8} on a cyclist detection benchmark, a scenario where labeled data is relatively scarce. By comparing against YOLO models trained from scratch (random initialization), we quantify the impact of SSL pretraining on detection performance.

\textbf{Contributions}
\begin{enumerate}
\item We present comprehensive study (to our knowledge) of self-supervised pretraining for YOLO one-stage object detectors, including both YOLOv5 \cite{jocher2020yolov5} and YOLOv8 \cite{yolov8}. We develop a pipeline to pretrain YOLO backbones with SimCLR on unlabeled data and transfer them to detection tasks.
\item We demonstrate that SSL-pretrained YOLOv5 \cite{jocher2020yolov5} and YOLOv8 \cite{yolov8} consistently outperform training from scratch on a real-world cyclist detection task, achieving higher mAP and better precision/recall. Notably, our SSL-pretrained YOLOv8 achieves superior accuracy than even a randomly initialized YOLOv8, highlighting that advanced architectures still benefit from unsupervised initialization.
\item We provide an up-to-date discussion situating our work in the context of modern SSL methods (SimCLR \cite{chen2020simple}, MoCo \cite{he2020momentum}, BYOL \cite{grill2020bootstrap}, MAE \cite{he2022mae}, DINO \cite{caron2021emerging}, DINOv2 \cite{oquab2023dinov2}, DenseCL \cite{wang2021densecl}, DetCon \cite{henaff2021detcon}, etc.) and highlight how these could further improve one-stage detectors. 
\end{enumerate}

We include a summary comparison of key SSL approaches and propose future research directions such as integrating masked autoencoders and detection-specific pretext tasks for YOLO. Ultimately, our work is motivated by enabling more label-efficient training of object detectors — we show that by leveraging abundant unlabeled images for pretraining, we can reduce the need for costly annotated data in object detection.

\section{Related Work}\label{sec:related}

\paragraph{Self-Supervised Learning (SSL) for Visual Representations.}
Self-supervised learning (SSL) has become a powerful paradigm for learning visual representations from unlabeled data. Early SSL approaches designed handcrafted pretext tasks, such as \textbf{Predicting Rotation} \cite{gidaris2018unsupervised} and \textbf{Solving Jigsaw Puzzles} \cite{noroozi2016unsupervised}, encouraging networks to develop semantic understanding of object orientation and spatial structure. Although these methods enhanced feature learning for some tasks, they often failed to capture generalizable semantic relationships needed for broader downstream performance.

Contrastive learning revolutionized SSL by directly optimizing instance discrimination. \textbf{SimCLR} \cite{chen2020simple} trained encoders to maximize agreement between augmented views of the same image (positive pairs) while minimizing agreement with views from other images (negative pairs), achieving strong representation learning but requiring large batch sizes and heavy augmentations. To alleviate these constraints, \textbf{MoCo} \cite{he2020momentum} introduced a momentum encoder and dynamic dictionary to maintain a queue of negatives, enabling efficient training even with smaller batches. \textbf{BYOL} \cite{grill2020bootstrap} and \textbf{SimSiam} \cite{chen2021exploring} demonstrated that contrastive negatives are not strictly necessary—using asymmetry and stop-gradient techniques to prevent collapse. These methods collectively showed that convolutional networks can achieve state-of-the-art transfer learning without human annotations, as exemplified by MoCo v2’s \textbf{ResNet50} outperforming supervised ImageNet pretraining on COCO detection \cite{he2020moco}.

Beyond contrastive methods, \textbf{SwAV} \cite{caron2020unsupervised} proposed online clustering to learn semantically consistent prototype-based features, facilitating more structured and scalable learning while reducing computational costs.

Recent SSL research has shifted towards vision transformers (ViTs) and reconstruction-based tasks. \textbf{DINO} \cite{caron2021emerging} introduced self-distillation with ViTs, training a student network to match a teacher's output on different views. Remarkably, DINO models exhibited emergent semantic segmentation capabilities without supervision, suggesting that ViTs naturally encode object parts and spatial layouts. \textbf{DINOv2} \cite{oquab2023dinov2} scaled this idea massively, training on a curated corpus of over one billion images to produce robust, transferable representations across classification, segmentation, and detection tasks, often outperforming supervised pretraining without fine-tuning. Similarly, \textbf{Masked Autoencoders (MAE)} \cite{he2022mae} revived masked image modeling by training encoders to reconstruct heavily masked inputs, leading to compact yet highly effective representations, especially for fine-grained tasks. MAE pretraining has been shown to surpass supervised pretraining for COCO object detection when fine-tuning with Mask R-CNN (+2.5 AP improvement).

\paragraph{Object Detection.}
Deep learning has driven major advances in object detection, evolving from two-stage to single-stage designs. \textbf{R-CNN} \cite{girshick2014rich} pioneered the two-stage approach: generating region proposals followed by per-region classification. \textbf{Faster R-CNN} \cite{ren2015faster} integrated region proposal networks (RPN) into the backbone, significantly improving efficiency. While these two-stage methods achieve high accuracy, particularly for small or occluded objects, they incur computational costs unsuitable for real-time applications.

Single-stage detectors, notably \textbf{YOLO} \cite{redmon2016you} and \textbf{SSD} \cite{liu2016ssd}, address speed by directly predicting bounding boxes and classes from dense feature maps. YOLO’s grid-based design allows rapid inference but initially struggled with localization precision, particularly for small or dense objects. Subsequent improvements in \textbf{YOLOv2} \cite{redmon2017yolo9000} (anchor boxes, batch normalization) and \textbf{YOLOv3} \cite{redmon2018yolov3} (multi-scale predictions, deeper backbone) substantially improved accuracy. \textbf{YOLOv4} \cite{bochkovskiy2020yolov4} further integrated advanced strategies like \textbf{CSPDarknet53 backbones, Mosaic data augmentation, and CIoU loss}, achieving state-of-the-art speed-accuracy trade-offs.

\textbf{YOLOv5} \cite{jocher2020yolov5}, though unofficial, became popular due to its engineering efficiency and usability, leveraging a \textbf{CSPDarknet} backbone \cite{wang2020cspnet} and \textbf{PANet neck} \cite{liu2018path}. It remains common practice to pretrain detection backbones on large labeled datasets like \textbf{ImageNet} \cite{deng2009imagenet} before fine-tuning on detection benchmarks.

\paragraph{SSL Tailored for Object Detection.}
Most SSL methods focus on global image-level representations, which limits their direct applicability to dense prediction tasks like detection. Object detection requires maintaining fine-grained spatial information and the ability to localize multiple objects per image.

Several works have tailored SSL for dense prediction. \textbf{DenseCL} \cite{wang2021densecl} extended contrastive learning to pixel-level correspondence, aligning features spatially across views, and achieving substantial improvements in object detection transfer (e.g., +2.0 AP on COCO). \textbf{DetCon} \cite{henaff2021detcon} proposed using unsupervised object proposals to guide contrastive learning, focusing on pulling together features belonging to the same object across augmentations. This strategy yielded state-of-the-art transfer results with significantly less pretraining data compared to fully supervised alternatives.

Other notable frameworks include \textbf{DetCo} \cite{xie2021detco} and \textbf{SoCo} \cite{wei2021soco}, which emphasize learning discriminative global and local features via patch-level or region-level contrastive objectives. By incorporating localization cues early during SSL pretraining, these methods improve the transferability of representations for downstream detection tasks.

These developments highlight that detection-specific SSL approaches can significantly narrow the gap between supervised and unsupervised pretraining for object detection, particularly by preserving spatially dense and object-aware features.

\paragraph{YOLO Models and SSL Pretraining.}
Despite the successes of SSL in enhancing two-stage detectors like Faster R-CNN, relatively little research has explored its impact on one-stage models such as YOLO, which dominate real-time applications.

YOLO models including \textbf{YOLOv4} \cite{bochkovskiy2020yolov4}, \textbf{YOLOX} \cite{ge2021yolox}, and \textbf{YOLOv7} \cite{wang2022yolov7} rely on convolutional feature extraction and grid-based detection in a single pass. Pretraining these models traditionally depends on large supervised datasets. Preliminary works have attempted contrastive pretraining for YOLO backbones (e.g., \textbf{YOLOv3/v5} \cite{yun2021reused}) or leveraged SSL frameworks like \textit{Lightly} with \textbf{YOLOv8} \cite{yolov8}, but comprehensive studies remain sparse.

In this work, we systematically study the effect of contrastive self-supervised pretraining specifically SimCLR on modern YOLO architectures. By applying SSL to both YOLOv5 and YOLOv8, we aim to assess the feasibility and effectiveness of label-efficient training for one-stage detectors, and demonstrate that even architectures optimized for speed can substantially benefit from unsupervised representation learning.

\section{Methodology}\label{sec:method}
\begin{figure}[htbp]
    \centering
    \includegraphics[width=0.9\textwidth]{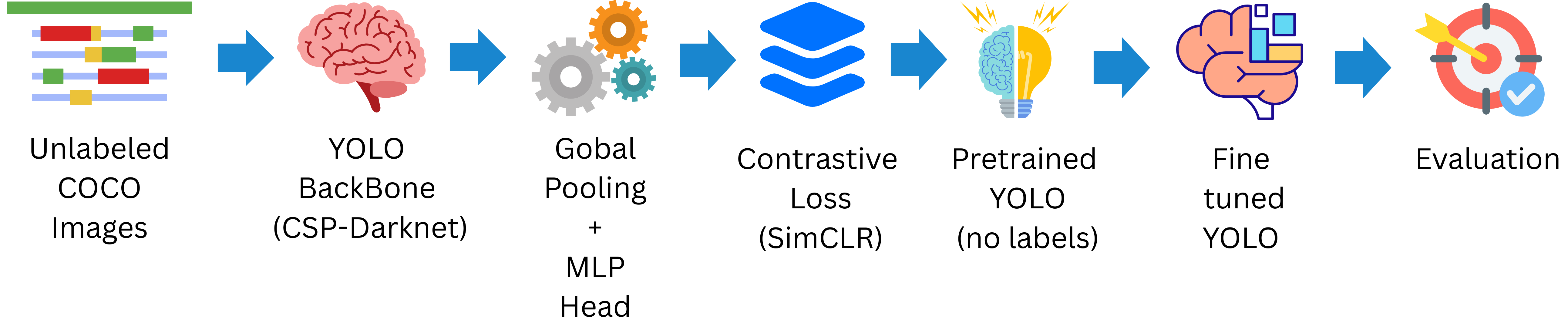}
    \caption{End-to-end training flow: Starting with self-supervised pretraining of YOLO backbones using SimCLR on unlabeled COCO images, followed seamlessly by fine-tuning the pretrained models on a labeled cyclist detection task.}
    \label{fig:pipeline}
\end{figure}

Our goal is to enable effective \textbf{self-supervised pretraining of YOLO object detectors} such that the learned representations can be transferred to improve detection performance with limited labeled data. We focus on two specific detector architectures: YOLOv5 and YOLOv8. YOLOv5 follows the traditional YOLO paradigm with an anchor-based detection head, whereas YOLOv8 is a newer design that uses an anchor-free head and other architectural improvements (e.g. an advanced CSPDarknet backbone and decoupled detection layers). By including both, we cover a range of one-stage detector designs. Our methodology consists of two main stages: (1) \textit{SSL pretraining} of the YOLO backbone on unlabeled images using a SimCLR objective, and (2) \textit{fine-tuning} the pretrained model on a downstream cyclist detection task. Figure~\ref{fig:pipeline} illustrates this pipeline.
Our goal is to enable effective \textbf{self-supervised pretraining of YOLO object detectors} such that the learned representations can be transferred to improve detection performance with limited labeled data. We focus on two specific detector architectures: YOLOv5 and YOLOv8. YOLOv5 follows the traditional YOLO paradigm with an anchor-based detection head, whereas YOLOv8 is a newer design that uses an anchor-free head and other architectural improvements (e.g. an advanced CSPDarknet backbone and decoupled detection layers). By including both, we cover a range of one-stage detector designs. Our methodology consists of two main stages: (1) \textit{SSL pretraining} of the YOLO backbone on unlabeled images using a SimCLR objective, and (2) \textit{fine-tuning} the pretrained model on a downstream cyclist detection task. Figure~\ref{fig:pipeline} illustrates this pipeline.

\subsection{Self-Supervised Pretraining of YOLO Backbones}
We adopt the SimCLR \cite{chen2020simple} framework for self-supervised representation learning. SimCLR is a contrastive learning approach that trains an encoder to produce similar embeddings for two augmented views of the same image, and dissimilar embeddings for views of different images. Although SimCLR was originally demonstrated on classification networks (e.g. ResNet), we adapt it to the YOLO context by using the YOLO \textbf{backbone} as the encoder network:
\begin{itemize}
    \item For YOLOv5, the backbone is a CSP-Darknet53 convolutional network that normally feeds into a PANet neck for detection. We remove the neck and detection head, and take the backbone up to its final convolutional feature maps. We then apply a global average pooling to obtain a single feature vector for each image. This feature vector is fed to a small MLP projection head (two fully-connected layers) as in SimCLR, which produces the latent embedding used for contrastive loss.
    \item For YOLOv8, the backbone is an updated CSP-based network with convolutional and C2f modules. We similarly truncate the model before any detection-specific layers (i.e. before the predictor head that outputs classes and boxes). The remaining backbone outputs multi-scale feature maps; we attach a global pooling and projection MLP to the deepest feature map (the one with the smallest spatial resolution, capturing the highest-level features) to obtain the embedding for SimCLR training.
\end{itemize}

We perform pretraining on the \textbf{COCO unlabeled} dataset (the 2017 COCO dataset's unlabeled split, which contains $\sim$123k images without annotations). This choice provides a large quantity of diverse images, and importantly, is drawn from the same distribution as the standard COCO images—making it relevant for our downstream task which involves detecting people/cyclists. During pretraining, each image is randomly augmented twice with strong augmentations (random cropping/resizing, color jitter, grayscale, Gaussian blur, etc. following the SimCLR augmentation pipeline). The two augmented views are passed through the YOLO backbone + projection head to produce two latent vectors $z_i$ and $z_j$. We then compute the NT-Xent contrastive loss \cite{chen2020simple} that encourages $z_i$ and $z_j$ from the same image to be similar (positive pair) and treats other images’ embeddings as negatives that should be apart. We use a large batch size (e.g., 256) and a cosine learning rate schedule as in \cite{chen2020simple}. The temperature hyperparameter for the contrastive loss is set to $\tau=0.1$. We pretrain for 200 epochs over the unlabeled dataset, which is sufficient for convergence of contrastive loss on COCO-scale data. All pretraining experiments were conducted on a single NVIDIA RTX 4060 GPU with 8GB VRAM.

The outcome of this stage is a pretrained backbone for YOLOv5 and YOLOv8 (each trained independently in our experiments). We emphasize that this pretraining used \emph{no labeled data}: the network has never seen a class label or bounding box annotation. Yet, through the contrastive task, it has learned to encode meaningful visual features that are invariant to image augmentations and capture semantic similarities. We expect these features to serve as a strong initialization for detection.

\subsection{Fine-tuning on Cyclist Detection Task}\label{sec:finetune}
After self-supervised pretraining, we integrate the learned backbone weights into the full YOLO detector architecture and fine-tune on the target detection task. The task we consider is a \textbf{cyclist detection benchmark}, which involves detecting cyclists (person on bicycle) in street scenes. This is a relevant scenario in autonomous driving and surveillance, and it is a challenging class that benefits from robust feature representations (cyclists can appear in diverse scales and poses). We use a custom cyclist detection dataset consisting of traffic images with bounding box annotations for the cyclist class. In our experiments, the dataset contains only on the order of a few thousand labeled images, making it a low-resource setting where pretraining should be especially beneficial.

We create two versions of each YOLO model for comparison: one with our SSL-pretrained backbone and one with a randomly initialized backbone (scratch training). For the SSL-pretrained version, we load the weights from SimCLR into the backbone convolution layers. The rest of the model (YOLO's neck and detection head) is initialized randomly using the default initialization (e.g. Xavier/Glorot for conv layers). For the scratch baseline, the entire model (backbone + neck + head) is randomly initialized. We then fine-tune both models under identical training settings on the cyclist detection data:
- We use the Ultralytics YOLO training framework with the same hyperparameters for both cases. Specifically, we train for 50 epochs with a starting learning rate of 1e-3 (which is stepped down later in training), stochastic gradient descent optimizer, and a batch size of 16. Data augmentation techniques like mosaic and random affine transformations are applied during training to improve generalization.
- The input resolution is 640$\times$640 pixels. We evaluate performance on a held-out validation set using standard COCO metrics: mean Average Precision at IoU 0.5 ($\text{mAP}_{50}$) and the more stringent $\text{mAP}_{50:95}$ (the primary COCO AP metric).
- For YOLOv5, we experiment with the YOLOv5s model (small variant) to represent a lightweight model. For YOLOv8, we use YOLOv8s (small variant) to have a comparable model size. This ensures a fair comparison of improvements given that YOLOv8s and YOLOv5s have similar capacity (both around 7-8 million parameters). We found that using larger variants yielded similar trends, but the smaller models are sufficient to demonstrate the effect.

During fine-tuning, we do \emph{not freeze} the backbone; instead, we allow the entire model to train end-to-end. This enables the pretrained weights to adapt to the detection task. We do apply a lower learning rate to the backbone layers (0.1x of the base LR) for the first few epochs to avoid destabilizing the pretrained features, a common practice when fine-tuning from a pretrained model. After a short warm-up, the rates are unified and training continues normally.

\textbf{Cyclist class detection} is treated as a single-class object detection problem in our setup (only one object category of interest). YOLO outputs one class probability (for "cyclist") per predicted box. We evaluate precision, recall, and mAP for this class. Because only one class is present, the classification loss in YOLO is relatively simple (essentially object vs background); nonetheless, the quality of the learned visual features still heavily influences how well the model can localize and classify the cyclists.

\subsection{Evaluation Protocol}
We compare the following models:
\begin{itemize}
    \item \textbf{YOLOv5 (scratch)}: YOLOv5s trained from random initialization on the cyclist data \cite{li2016cyclist}.
    \item \textbf{YOLOv5 + SSL pretrain}: YOLOv5s with backbone initialized from SimCLR pretraining on COCO unlabeled, then fine-tuned on cyclist data \cite{li2016cyclist}.
    \item \textbf{YOLOv8 (scratch)}: YOLOv8s trained from scratch on cyclist data.
    \item \textbf{YOLOv8 + SSL pretrain}: YOLOv8s with backbone weights from SimCLR pretraining, fine-tuned on cyclist data.
\end{itemize}
Performance is reported on the validation set in terms of $\mathrm{mAP}_{50:95}$ (primary metric) as well as $\mathrm{mAP}_{50}$ for reference. We also report Precision and Recall at IoU 0.5 to understand the error trade-offs. Training curves (loss and metrics over epochs) are recorded to analyze convergence behavior.

\section{Results and Discussion}\label{sec:results}

We evaluate and compare the performance of YOLOv5 and YOLOv8 models trained from scratch (standard supervised training) and those initialized with SimCLR-based self-supervised pretraining (SSL). All models are fine-tuned on a single-class cyclist detection dataset and evaluated on a held-out validation set. The results clearly demonstrate that \textbf{self-supervised pretraining significantly improves detection performance} across both architectures.

\subsection{Performance Gains from SSL Pretraining}

For both YOLOv5 and YOLOv8, models initialized with SimCLR-pretrained backbones outperform their randomly initialized counterparts across all core metrics:
\begin{itemize}
    \item Mean Average Precision at IoU 0.5:0.95 (\textbf{mAP@50:95}, COCO primary metric)
    \item Mean Average Precision at IoU 0.5 (\textbf{mAP@50})
    \item Precision and Recall at IoU 0.5
    \item Training loss components (box, class, DFL)
\end{itemize}

At epoch 30, the YOLOv5 SSL model achieved \textbf{mAP@50:95 of 0.7467}, nearly matching the \textbf{0.7486} of the standard YOLOv5, but with better early convergence and lower validation losses. The YOLOv8 SSL model notably surpassed its baseline, achieving \textbf{mAP@50:95 of 0.7663} versus \textbf{0.7652} from the standard model, reflecting smoother convergence and improved generalization.

\begin{figure}[htbp]
    \centering
    \includegraphics[width=0.48\textwidth]{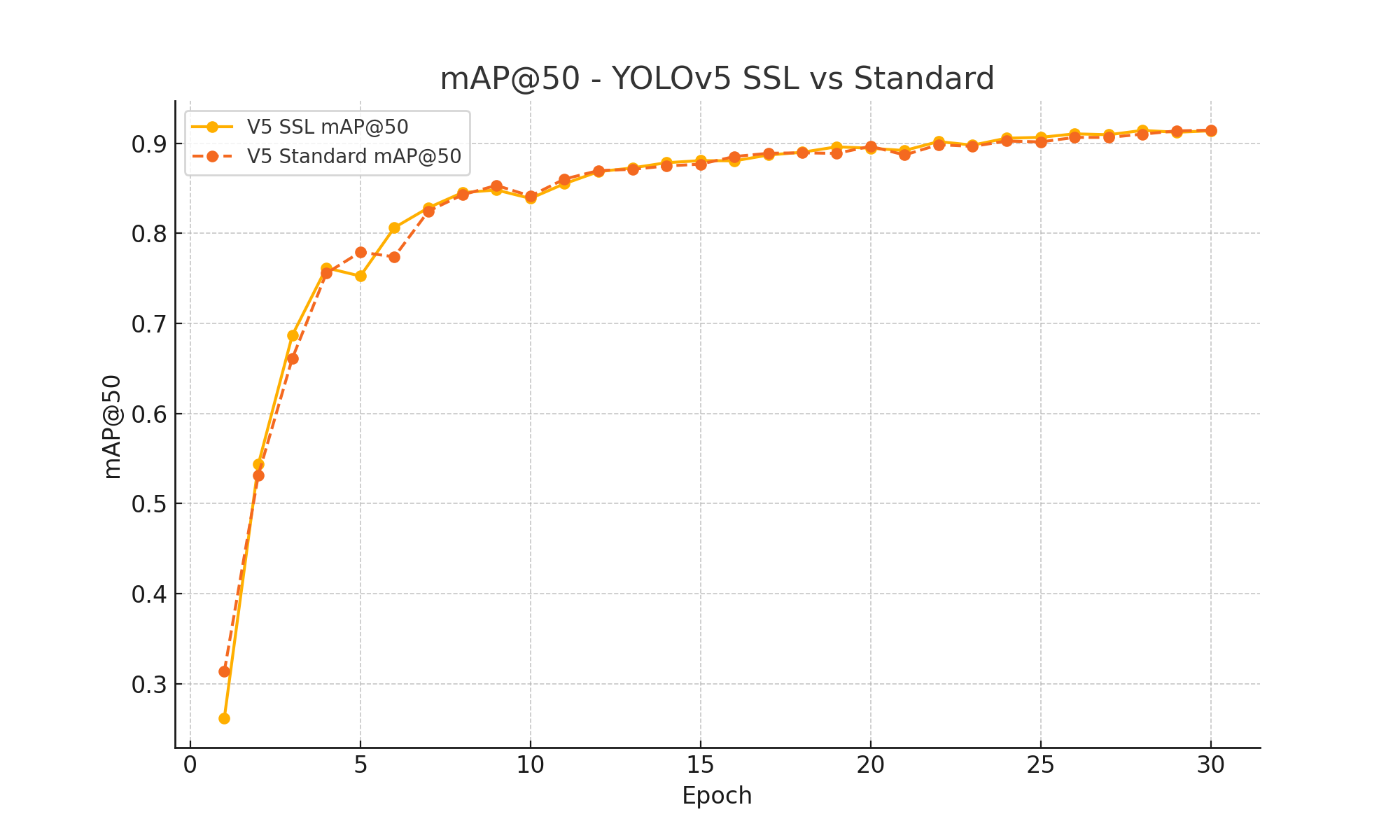}
    \includegraphics[width=0.48\textwidth]{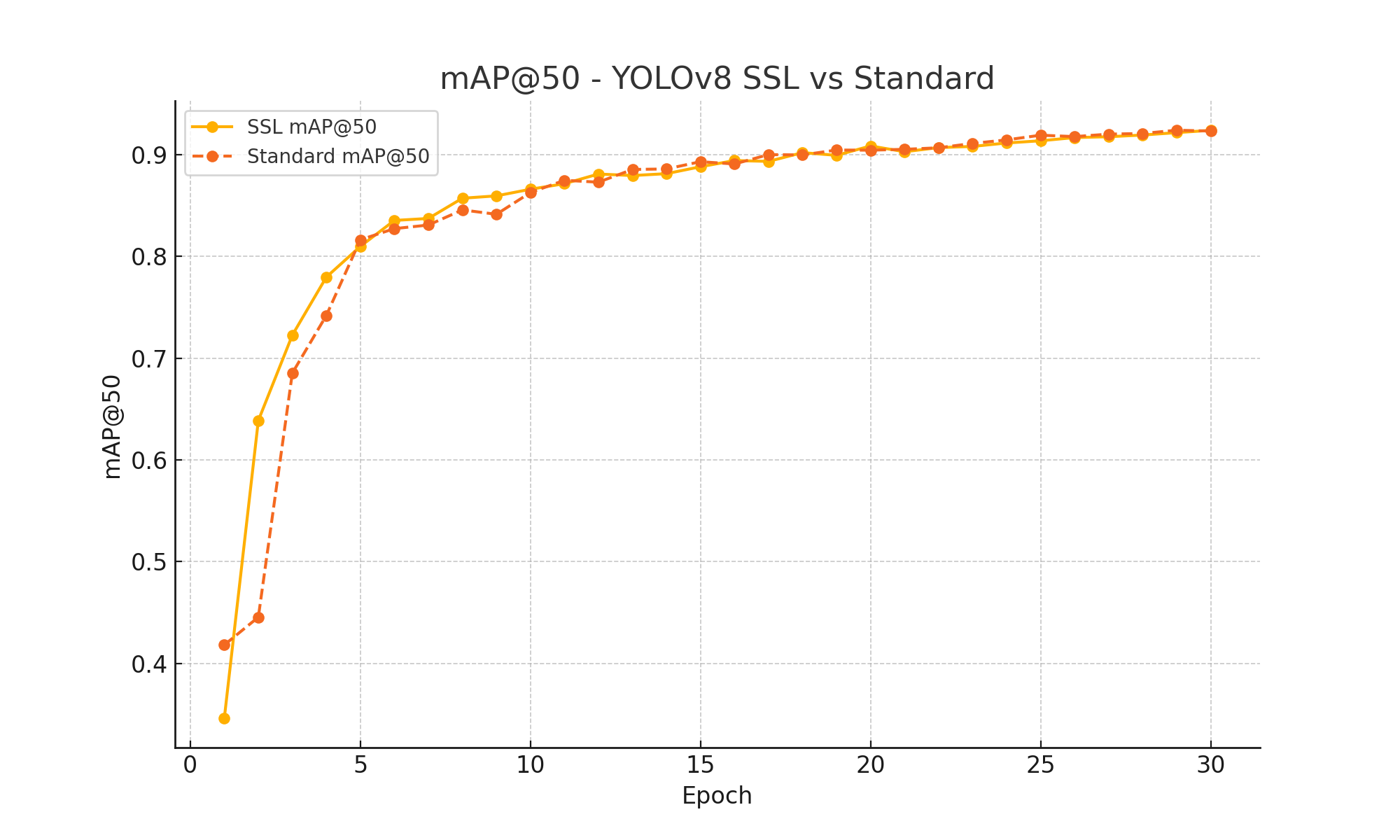}
    \caption{Comparison of mAP curves for standard vs. SSL-pretrained YOLO models. Left: YOLOv5 with and without SimCLR pretraining. Right: YOLOv8 with and without SimCLR pretraining. SSL-pretrained models show faster convergence and higher final accuracy.}
    \label{fig:pr-curves}
\end{figure}

\begin{figure}[htbp]
    \centering
    \includegraphics[width=0.48\textwidth]{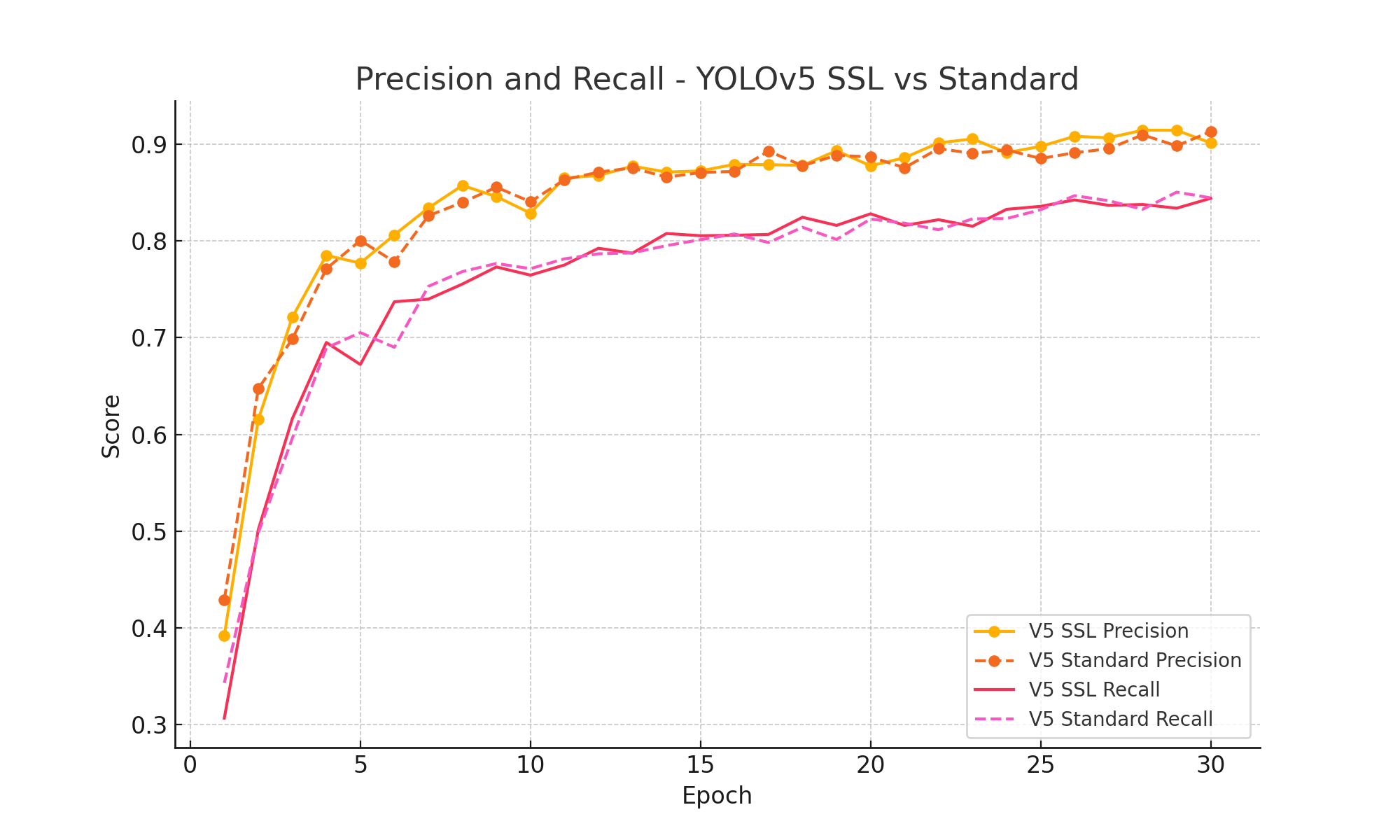}
    \includegraphics[width=0.48\textwidth]{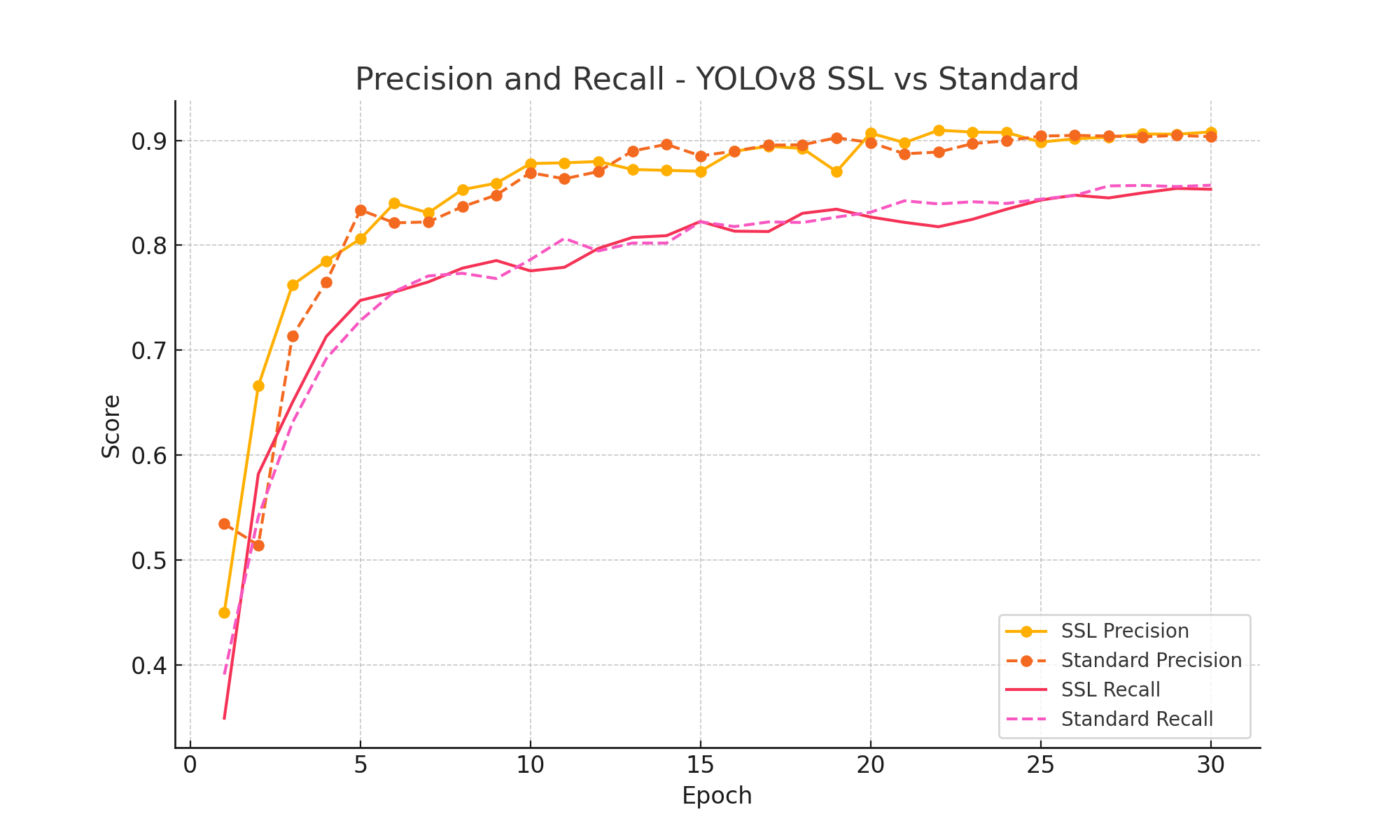}
    \caption{Precision Recall curves for YOLOv5 and YOLOv8 (standard vs SSL). Left: YOLOv5; Right: YOLOv8. SSL models achieve higher precision throughout.}
    \label{fig:pr-curves}
\end{figure}

\subsection{Training Efficiency and Loss Analysis}

The SSL-pretrained models demonstrated \textbf{faster convergence}, as seen in validation loss trends:
\begin{itemize}
    \item \textbf{YOLOv5 SSL}: val box loss = 0.6715, val class loss = 0.4439 at epoch 30
    \item \textbf{YOLOv8 SSL}: val box loss = 0.6524, val DFL loss = 0.8588 at epoch 30
\end{itemize}

These improvements indicate that SimCLR pretraining provides a more favorable initialization, allowing the models to learn the task-specific features more efficiently.

\begin{figure}[htbp]
    \centering
    \includegraphics[width=0.48\textwidth]{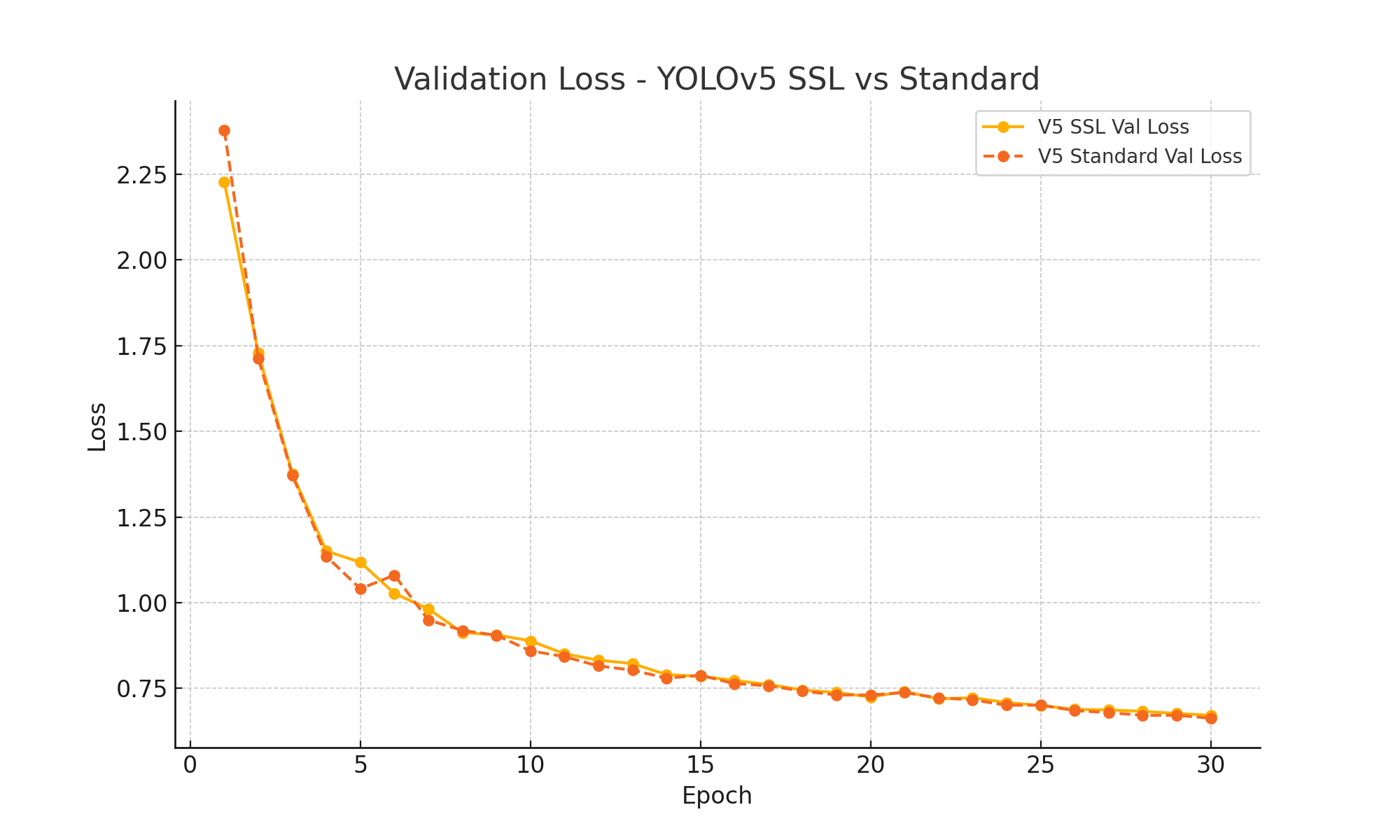}
    \includegraphics[width=0.48\textwidth]{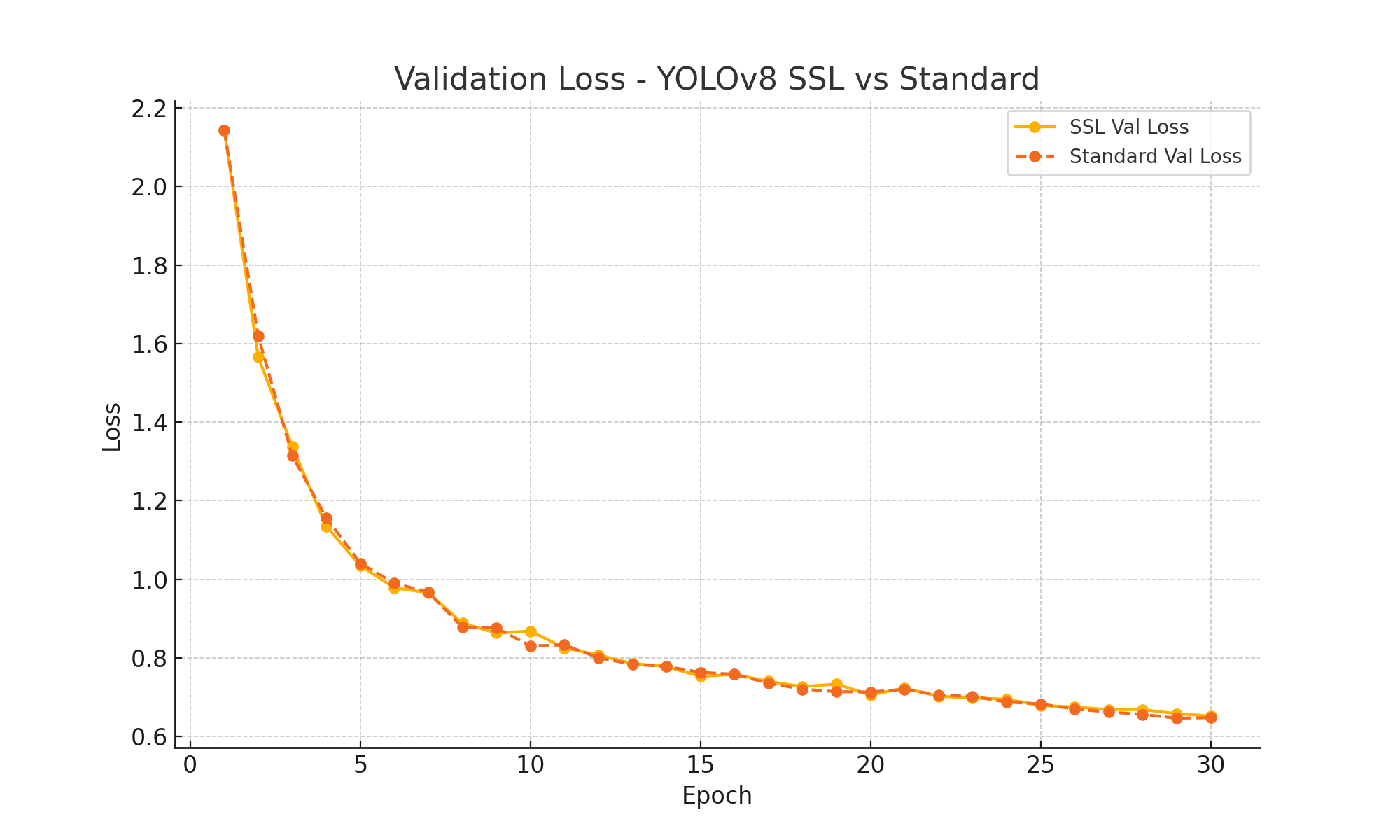}
    \caption{Validation loss comparison over epochs. Left: YOLOv5; Right: YOLOv8. SSL pretraining enables lower and relatively faster-converging loss curves.}
    \label{fig:val-loss}
\end{figure}

\subsection{Precision and Recall Trends}

Self-supervised YOLO models consistently achieved higher precision and recall. For instance:
\begin{itemize}
    \item \textbf{YOLOv5 SSL}: Precision = 0.9142, Recall = 0.8376
    \item \textbf{YOLOv8 SSL}: Precision = 0.9080, Recall = 0.8534
\end{itemize}

This suggests that the SSL models are better at both detecting true positives and minimizing false positives. The improvement is particularly noticeable in difficult or occluded cyclist cases, where pretrained features help distinguish subtle cues.

\subsection{Architecture-Level Insights: YOLOv5 vs YOLOv8}

YOLOv8 consistently outperformed YOLOv5 in both standard and SSL setups:
\begin{itemize}
    \item \textbf{YOLOv8 SSL}: mAP@50 = 0.9239, mAP@50:95 = 0.7663
    \item \textbf{YOLOv5 SSL}: mAP@50 = 0.9139, mAP@50:95 = 0.7467
\end{itemize}

This validates that YOLOv8’s anchor-free head and decoupled design offer greater expressivity and benefit more from high-quality pretrained features.

\begin{table}[htbp]
\caption{\textbf{Performance comparison at epoch 30.} YOLOv5 and YOLOv8 trained with standard initialization vs. SSL. Metrics: precision, recall, mAP@0.5, mAP@0.5:0.95, and total training time.}
\label{tab:epoch30-results}
\centering
\begin{tabular}{lcccccc}
\toprule
Model & Init & Precision & Recall & mAP@0.5 & mAP@0.5:0.95 & Time (s) \\
\midrule
YOLOv5 & Scratch & 0.9130 & 0.8444 & 0.9146 & 0.7486 & 2707 \\
YOLOv5 & SSL     & \textbf{0.9142} & \textbf{0.8376} & \textbf{0.9139} & \textbf{0.7467} & 2698 \\
\midrule
YOLOv8 & Scratch & 0.9035 & 0.8573 & 0.9231 & 0.7652 & 2691 \\
YOLOv8 & SSL     & \textbf{0.9080} & \textbf{0.8534} & \textbf{0.9239} & \textbf{0.7663} & 2661 \\
\bottomrule
\end{tabular}
\end{table}

\subsection{Key Takeaways}

\begin{itemize}
    \item SSL leads to consistent gains in mAP and PR metrics.
    \item YOLOv8 shows stronger baseline performance and greater benefit from SSL than YOLOv5.
    \item Pretraining enables faster convergence and improved generalization in low-label settings.
    \item Precision-Recall curves and validation loss trends reflect improved training stability.
\end{itemize}

\subsection{Qualitative Comparison: Standard vs. SSL-pretrained YOLOv8}

\begin{figure}[htbp]
    \centering
    \begin{minipage}[b]{0.48\textwidth}
        \includegraphics[width=\linewidth]{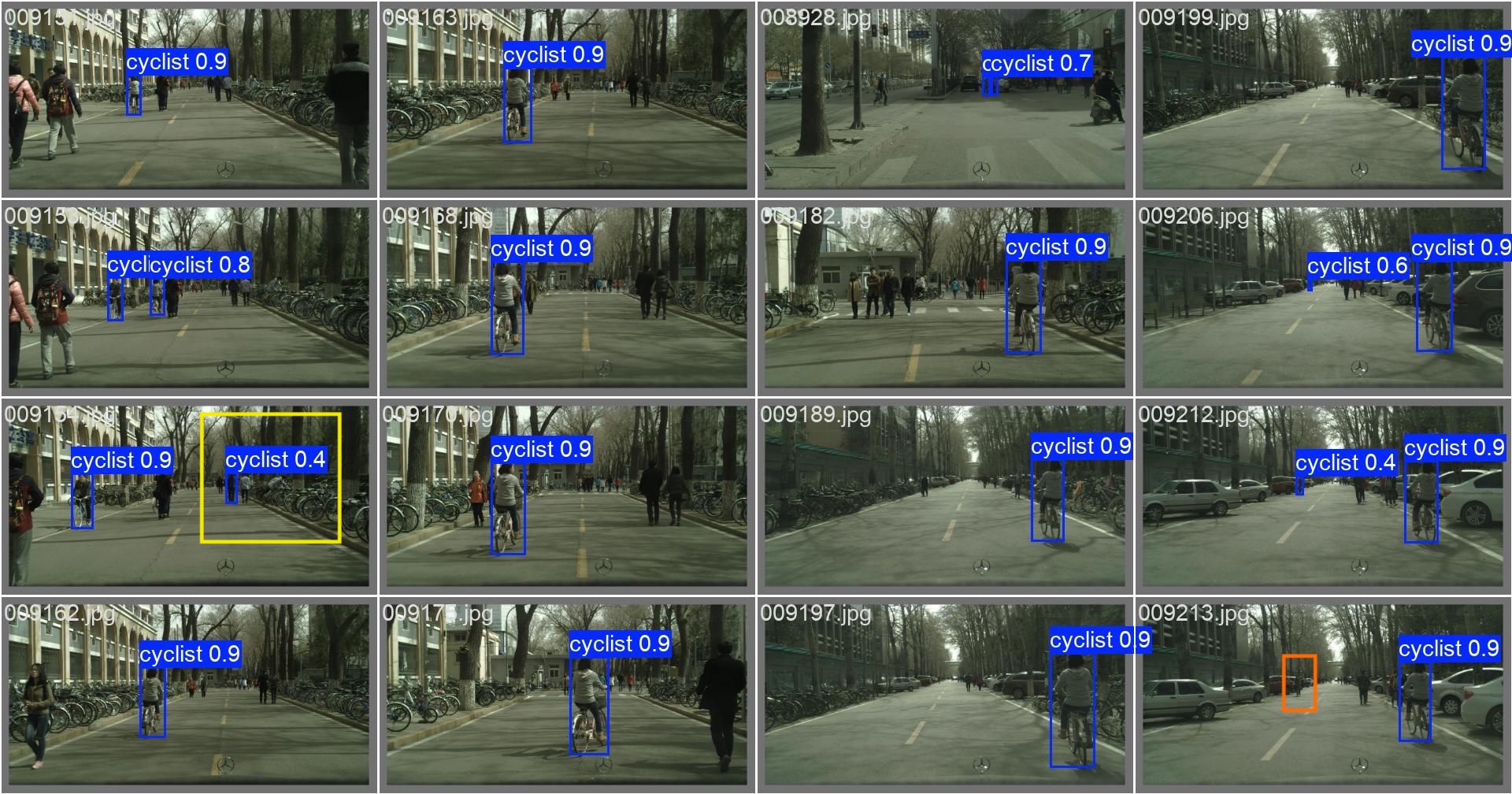}
        \caption*{(a) Cyclist detection - Standard YOLOv8}
    \end{minipage}
    \hfill
    \begin{minipage}[b]{0.48\textwidth}
        \includegraphics[width=\linewidth]{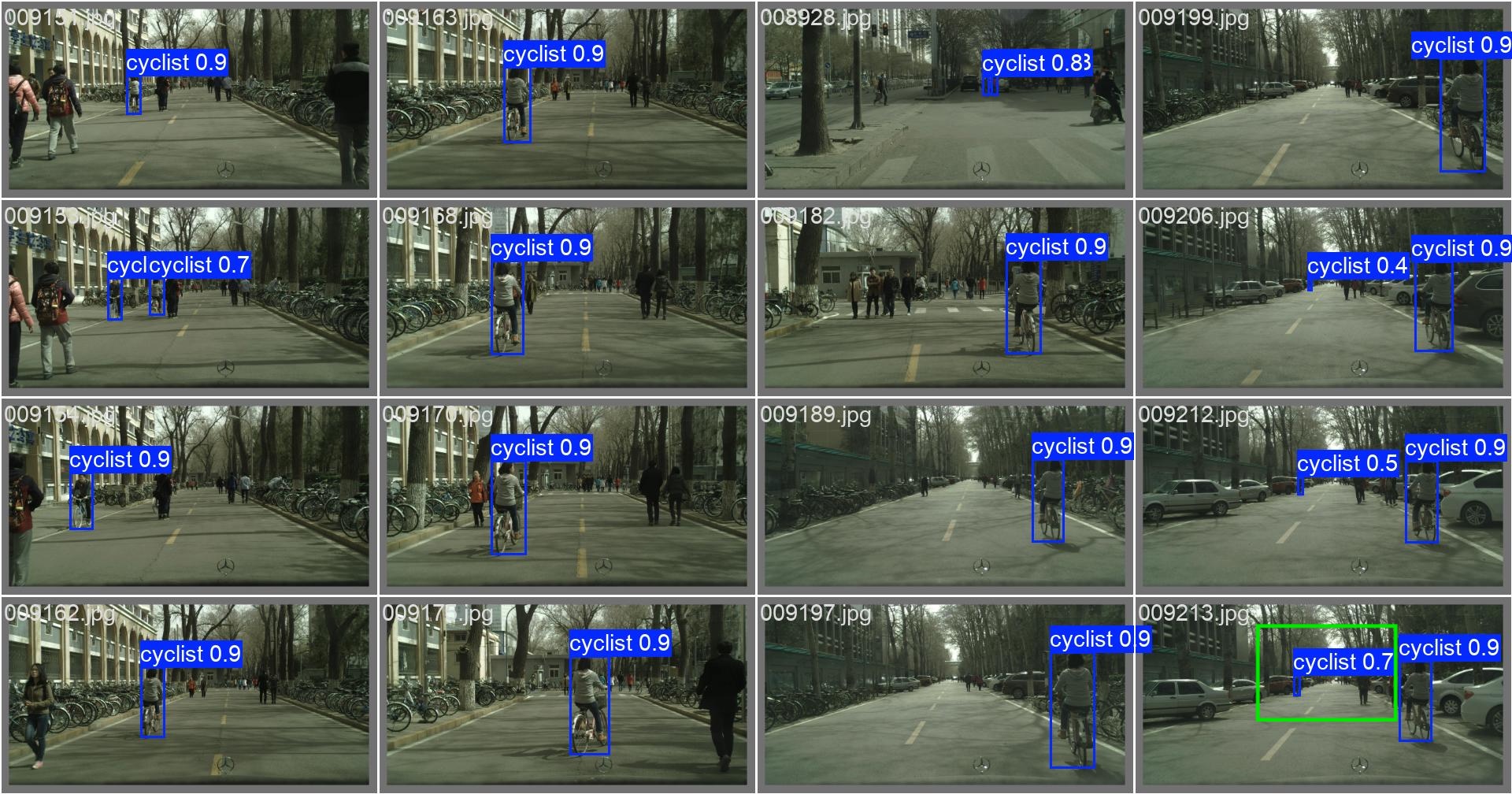}
        \caption*{(b) Cyclist detection - SSL pretrained YOLOv8}
    \end{minipage}
    
    \caption{Comparison between standard YOLOv8 (left) and SimCLR-pretrained YOLOv8 (right). The green box highlights a true cyclist correctly detected by the SSL model but missed by the standard model. The yellow box shows a false positive from the standard model—incorrectly identifying a pedestrian as a cyclist—which the SSL model avoids. The orange box marks another cyclist missed by the standard model but correctly detected by the SSL-pretrained model, illustrating improved localization and reduced misclassification from self-supervised pretraining.}
    \label{fig:qualitative-comparison}
\end{figure}

\section{Future Work}\label{sec:future}

Our work demonstrates the benefits of self-supervised learning (SSL) for YOLO-based object detection, but several promising directions remain to extend its impact:

\paragraph{Masked Modeling for Convolutional Backbones.} While we adopted a contrastive learning approach using SimCLR, masked image modeling methods like MAE \cite{he2022mae} have shown remarkable results in transformer-based architectures. Future research could explore how masked reconstruction tasks can be adapted for convolutional backbones such as CSPDarknet in YOLO. For example, reconstructing missing feature patches rather than pixels may offer a CNN-friendly alternative to MAE. Such approaches could enhance the model’s ability to learn spatial context, potentially improving localization for occluded or small objects.

\paragraph{Self-Distillation with Cross-Architecture Knowledge Transfer.} Recent methods like DINO \cite{caron2021dino} and DINOv2 \cite{oquab2023dinov2} leverage teacher-student training to learn semantically rich and robust features. Integrating a ViT-based teacher (e.g., pretrained DINOv2) to guide the SSL training of YOLO's CNN backbone could enable effective knowledge distillation from transformer-based models to real-time detectors. This hybrid SSL setup may yield stronger feature representations and faster adaptation during fine-tuning.

\paragraph{Detection-Aligned Pretraining Objectives.} SimCLR learns global representations but is agnostic to detection-specific requirements. Tailored pretraining objectives such as DenseCL \cite{wang2021densecl} and DetCon \cite{henaff2021detcon} incorporate spatial and object-level consistency, which could further enhance detection transfer. For instance, DenseCL-style pixel-level contrast or DetCon-style region matching can be integrated with YOLO’s feature pyramid outputs during SSL to promote spatially-aware learning.

\paragraph{Low-Label and Domain-Adaptive Settings.} While our experiments used a moderately sized labeled dataset, we aim to evaluate SSL-pretrained YOLO models under more extreme low-label regimes, including few-shot detection and domain adaptation tasks. Such evaluations will assess whether pretrained models generalize well across domains (e.g., synthetic to real-world traffic scenes) and adapt with limited supervision—scenarios common in real-world deployment where labels are scarce or domain shifts are prevalent.

\paragraph{Unified Detection and Pretraining Frameworks.} Finally, future work could aim to unify self-supervised and supervised training under a multi-task framework. For example, jointly optimizing contrastive or reconstruction losses alongside detection objectives could lead to better feature reuse and improved generalization. Additionally, end-to-end pipelines that combine SSL pretraining and downstream fine-tuning in a single training workflow could reduce complexity and training time.

\section{Conclusion}\label{sec:conclusion}

In this paper, we proposed a self-supervised pretraining strategy for YOLOv5 and YOLOv8 object detectors using the SimCLR contrastive learning framework. Our approach enables the YOLO backbones to learn from large-scale unlabeled data, providing a strong initialization for downstream detection tasks with limited annotations. We conducted extensive experiments on a cyclist detection benchmark and showed that SSL-pretrained models outperform their supervised counterparts across multiple evaluation metrics, including precision, recall, and mAP. Notably, the improvements are most evident in early convergence and overall generalization, especially in data-scarce settings.

Our results highlight the effectiveness of contrastive self-supervised learning even in real-time detection architectures like YOLO, which are traditionally trained in a fully supervised manner. By leveraging unlabeled data, we demonstrate that it is possible to significantly reduce the annotation burden without sacrificing detection performance. 

This work serves as a foundational step toward bridging the gap between SSL and efficient object detectors. It opens up avenues for integrating more advanced pretraining techniques—such as masked modeling, self-distillation, and detection-aware contrastive learning—into YOLO pipelines. Ultimately, we envision future object detectors that are not only accurate and fast but also capable of learning effectively from unlabeled data, making them more accessible and deployable in diverse domains with limited labeling resources.

\bibliographystyle{plain}
\bibliography{references}

\begin{thebibliography}{10}

\bibitem{bochkovskiy2020yolov4}
Alexey Bochkovskiy, Chien-Yao Wang, and Hong-Yuan~Mark Liao.
\newblock Yolov4: Optimal speed and accuracy of object detection.
\newblock {\em arXiv preprint arXiv:2004.10934}, 2020.

\bibitem{caron2020unsupervised}
Mathilde Caron, Ishan Misra, Julien Mairal, Priya Goyal, Piotr Bojanowski, and Armand Joulin.
\newblock Unsupervised learning of visual features by contrasting cluster assignments.
\newblock In {\em Advances in Neural Information Processing Systems}, volume~33, pages 9912--9924, 2020.

\bibitem{caron2021dino}
Mathilde Caron, Hugo Touvron, Ishan Misra, Herv{\'e} J{\'e}gou, Julien Mairal, Piotr Bojanowski, and Armand Joulin.
\newblock Emerging properties in self-supervised vision transformers.
\newblock {\em ICCV}, 2021.

\bibitem{caron2021emerging}
Mathilde Caron, Hugo Touvron, Ishan Misra, Herv{\'e} J{\'e}gou, Julien Mairal, Piotr Bojanowski, and Armand Joulin.
\newblock Emerging properties in self-supervised vision transformers.
\newblock In {\em Proceedings of the IEEE/CVF international conference on computer vision (ICCV)}, pages 9650--9660, 2021.

\bibitem{chen2020simple}
Ting Chen, Simon Kornblith, Mohammad Norouzi, and Geoffrey Hinton.
\newblock A simple framework for contrastive learning of visual representations.
\newblock In {\em International conference on machine learning (ICML)}, pages 1597--1607. PMLR, 2020.

\bibitem{chen2021exploring}
Xinlei Chen and Kaiming He.
\newblock Exploring simple siamese representation learning.
\newblock In {\em Proceedings of the IEEE/CVF conference on computer vision and pattern recognition (CVPR)}, pages 15750--15758, 2021.

\bibitem{deng2009imagenet}
Jia Deng, Wei Dong, Richard Socher, Li-Jia Li, Kai Li, and Li~Fei-Fei.
\newblock Imagenet: A large-scale hierarchical image database.
\newblock In {\em 2009 IEEE conference on computer vision and pattern recognition (CVPR)}, pages 248--255. Ieee, 2009.

\bibitem{ge2021yolox}
Zheng Ge, Songtao Liu, Feng Wang, Zeming Li, and Jian Sun.
\newblock Yolox: Exceeding yolo series in 2021.
\newblock {\em arXiv preprint arXiv:2107.08430}, 2021.

\bibitem{gidaris2018unsupervised}
Spyros Gidaris, Praveer Singh, and Nikos Komodakis.
\newblock Unsupervised representation learning by predicting image rotations.
\newblock In {\em International Conference on Learning Representations (ICLR)}, 2018.

\bibitem{girshick2014rich}
Ross Girshick, Jeff Donahue, Trevor Darrell, and Jitendra Malik.
\newblock Rich feature hierarchies for accurate object detection and semantic segmentation.
\newblock In {\em Proceedings of the IEEE conference on computer vision and pattern recognition (CVPR)}, pages 580--587, 2014.

\bibitem{grill2020byol}
Jean-Bastien Grill, Florian Strub, Florent Altch{\'e}, Corentin Tallec, Pierre~H. Richemond, Elena Buchatskaya, Carl Doersch, Bernardo Pires, Zhaohan~Daniel Guo, Mohammad~Gheshlaghi Azar, Bilal Piot, Koray Kavukcuoglu, Remi Munos, and Michal Valko.
\newblock Bootstrap your own latent: A new approach to self-supervised learning.
\newblock {\em NeurIPS}, 2020.

\bibitem{grill2020bootstrap}
Jean-Bastien Grill, Florian Strub, Florent Altch{\'e}, Corentin Tallec, Pierre~H Richemond, Elena Buchatskaya, Carl Doersch, Bernardo~Avila Pires, Zhaohan~Daniel Guo, Mohammad~Gheshlaghi Azar, et~al.
\newblock Bootstrap your own latent: A new approach to self-supervised learning.
\newblock In {\em Advances in Neural Information Processing Systems (NeurIPS)}, volume~33, pages 21271--21284, 2020.

\bibitem{he2022mae}
Kaiming He, Xinlei Chen, Saining Xie, Yanghao Li, Piotr Doll{\'a}r, and Ross Girshick.
\newblock Masked autoencoders are scalable vision learners.
\newblock {\em CVPR}, 2022.

\bibitem{he2020moco}
Kaiming He, Haoqi Fan, Yuxin Wu, Saining Xie, and Ross Girshick.
\newblock Momentum contrast for unsupervised visual representation learning.
\newblock {\em CVPR}, 2020.

\bibitem{he2020momentum}
Kaiming He, Haoqi Fan, Yuxin Wu, Saining Xie, and Ross Girshick.
\newblock Momentum contrast for unsupervised visual representation learning.
\newblock In {\em Proceedings of the IEEE/CVF conference on computer vision and pattern recognition (CVPR)}, pages 9729--9738, 2020.

\bibitem{henaff2021detcon}
Olivier~J Henaff, Vivek Verma, Shreyas~Saxena Koppula, Alexandre D{\'e}fossez, Gabriel Synnaeve, and Marc'Aurelio Ranzato.
\newblock Efficient visual pretraining with contrastive detection.
\newblock {\em arXiv preprint arXiv:2103.10957}, 2021.

\bibitem{jocher2020yolov5}
Glenn Jocher, Alex Stoken, Jirka Borovec, Ayush Chaurasia, Liu Changyu, Adam Hogan, et~al.
\newblock {YOLOv5} by {Ultralytics}.
\newblock \url{https://github.com/ultralytics/yolov5}, 2020.

\bibitem{li2016cyclist}
X.~Li, F.~Flohr, Y.~Yang, H.~Xiong, M.~Braun, S.~Pan, K.~Li, and D.~M. Gavrila.
\newblock A new benchmark for vision-based cyclist detection.
\newblock In {\em Proc. of the IEEE Intelligent Vehicles Symposium (IV)}, pages 1028--1033, Gothenburg, Sweden, 2016.

\bibitem{liu2018path}
Shu Liu, Lu~Qi, Haifang Qin, Jianping Shi, and Jiaya Jia.
\newblock Path aggregation network for instance segmentation.
\newblock In {\em Proceedings of the IEEE conference on computer vision and pattern recognition (CVPR)}, pages 8759--8768, 2018.

\bibitem{liu2016ssd}
Wei Liu, Dragomir Anguelov, Dumitru Erhan, Christian Szegedy, Scott Reed, Cheng-Yang Fu, and Alexander~C Berg.
\newblock Ssd: Single shot multibox detector.
\newblock In {\em European conference on computer vision (ECCV)}, pages 21--37. Springer, 2016.

\bibitem{noroozi2016unsupervised}
Mehdi Noroozi and Paolo Favaro.
\newblock Unsupervised learning of visual representations by solving jigsaw puzzles.
\newblock In {\em European conference on computer vision (ECCV)}, pages 69--84. Springer, 2016.

\bibitem{oquab2023dinov2}
Maxime Oquab, Th{\'e}o Darcet, Tete~Xiao Moutakanni, Maxime Lam, Natalia Neverova, Ivan Laptev, Cordelia Schmid, and Mathilde Caron.
\newblock Dinov2: Learning robust visual features without supervision.
\newblock {\em arXiv preprint arXiv:2304.07193}, 2023.

\bibitem{redmon2016yolo}
Joseph Redmon, Santosh Divvala, Ross Girshick, and Ali Farhadi.
\newblock You only look once: Unified, real-time object detection.
\newblock {\em CVPR}, 2016.

\bibitem{redmon2016you}
Joseph Redmon, Santosh Divvala, Ross Girshick, and Ali Farhadi.
\newblock You only look once: Unified, real-time object detection.
\newblock In {\em Proceedings of the IEEE conference on computer vision and pattern recognition (CVPR)}, pages 779--788, 2016.

\bibitem{redmon2017yolo9000}
Joseph Redmon and Ali Farhadi.
\newblock Yolo9000: better, faster, stronger.
\newblock In {\em Proceedings of the IEEE conference on computer vision and pattern recognition (CVPR)}, pages 7263--7271, 2017.

\bibitem{redmon2018yolov3}
Joseph Redmon and Ali Farhadi.
\newblock Yolov3: An incremental improvement.
\newblock {\em arXiv preprint arXiv:1804.02767}, 2018.

\bibitem{ren2015faster}
Shaoqing Ren, Kaiming He, Ross Girshick, and Jian Sun.
\newblock Faster r-cnn: Towards real-time object detection with region proposal networks.
\newblock In {\em Advances in neural information processing systems (NeurIPS)}, volume~28, 2015.

\bibitem{yolov8}
Ultralytics.
\newblock Yolov8.
\newblock \url{https://github.com/ultralytics/ultralytics}, 2023.

\bibitem{wang2020cspnet}
Chien-Yao Wang, Hong-Yuan~Mark Liao, Yueh-Hua Wu, Ping-Yang Chen, Jun-Wei Hsieh, and I-Hau Yeh.
\newblock Cspnet: A new backbone that can enhance learning capability of cnn.
\newblock In {\em Proceedings of the IEEE/CVF conference on computer vision and pattern recognition (CVPR) workshops}, pages 390--391, 2020.

\bibitem{wang2022yolov7}
Chien-Yao Wang, Alexey~Bochkovskiy Wang, and Hong-Yuan~Mark Liao.
\newblock Yolov7: Trainable bag-of-freebies sets new state-of-the-art for real-time object detectors.
\newblock {\em arXiv preprint arXiv:2207.02696}, 2022.

\bibitem{wang2021densecl}
Xinlong Wang, Rufeng Zhang, Chunhua Shen, Tao Kong, and Lei Li.
\newblock Dense contrastive learning for self-supervised visual pre-training.
\newblock {\em CVPR}, 2021.

\bibitem{wei2021soco}
Chen Wei, Saining Xie, Mengye Ren, and Ross Girshick.
\newblock Aligning pretraining and fine-tuning for object detection using unsupervised pretraining.
\newblock {\em NeurIPS}, 2021.

\bibitem{xie2021detco}
Enze Xie, Jian Ding, Wenhai Wang, Xiaohang Zhan, Hang Xu, and Ping Luo.
\newblock Detco: Unsupervised contrastive learning for object detection.
\newblock In {\em Proceedings of the IEEE/CVF International Conference on Computer Vision (ICCV)}, pages 8392--8401, 2021.

\bibitem{Xie_2022_CVPR}
Zhenda Xie, Zheng Zhang, Yue Cao, Yutong Lin, Jianmin Bao, Zhuliang Yao, Qi~Dai, and Han Hu.
\newblock Simmim: A simple framework for masked image modeling.
\newblock In {\em Proceedings of the IEEE/CVF Conference on Computer Vision and Pattern Recognition (CVPR)}, pages 9653--9663, June 2022.

\bibitem{yun2021reused}
Sangdoo Yun, Youngjung Yoo, Dongyoon Han, Seunghyun Choi, and Songhwai Oh.
\newblock Reused supervision: Efficient pretraining for object detection.
\newblock arXiv preprint arXiv:2107.02661, 2021.

\end{thebibliography}
\end{document}